\pdfoutput=1

\documentclass[11pt]{article}

\usepackage[]{acl}

\usepackage{times}
\usepackage{latexsym}

\usepackage[T1]{fontenc}

\usepackage[utf8]{inputenc}

\usepackage{microtype}

\usepackage{inconsolata}

\usepackage{graphicx}
\usepackage{enumitem}

\newcommand{\turnwise}{\textsc{TurnWise}\xspace}
\newcommand{\dataset}{\textsc{TurnWiseData}\xspace}
\newcommand{\datasetEval}{\textsc{TurnWiseEval}\xspace}

\usepackage{colortbl, xcolor, booktabs, pifont}

\usepackage[table,xcdraw]{xcolor}
\usepackage{xspace}
\usepackage{graphicx}

\usepackage[most]{tcolorbox}

\definecolor{darkgreen}{rgb}{0.0, 0.8, 0.0}

\newcolumntype{x}[1]{>{\centering\arraybackslash\hspace{0pt}}p{#1}}

\title{\turnwise: The Gap between \\ Single- and Multi-turn Language Model Capabilities}

\author{
Victoria Graf$^{1,2}$ \And
Valentina Pyatkin$^{2}$ \And
Nouha Dziri$^{2}$ \\
$^{1}$University of Washington \quad
$^{2}$Allen Institute for AI \And
\hspace{-25pt} Nathan Lambert$^{2}$ \And
Hannaneh Hajishirzi$^{1,2}$
}

\begin{document}
\maketitle
\begin{abstract}
  Multi-turn conversations are a common and critical mode of language model interaction.
  However, current open training and evaluation data focus on single-turn settings, failing to capture the additional dimension of these longer interactions.
  To understand this multi-/single-turn gap, we first introduce a new benchmark, \datasetEval, for multi-turn capabilities that is directly comparable to single-turn chat evaluation.
  Our evaluation isolates multi-turn specific conversational ability through pairwise comparison to equivalent single-turn settings.
  We additionally introduce our synthetic multi-turn data pipeline \dataset which allows the scalable generation of multi-turn training data. Our experiments with Olmo 3 show that training with multi-turn data is vital to achieving strong multi-turn chat performance, and that including as little as 10k multi-turn conversations during post-training can lead to a 12\% improvement on \datasetEval.
\end{abstract}

\section{Introduction}

Large language models (LLMs) are increasingly deployed to real-world users for free-form, multi-turn conversations. However, many training and evaluation methodologies continue to rely primarily on single-turn data, creating a mismatch between real-world usage and current paradigms.
Open language modeling recipes \cite{tulu3, olmo3} focus on and are primarily comprised of single-turn data. Where multi-turn data is included in these recipes, it is added without systematic consideration of its contribution to conversational abilities.
Moreover, collecting scalable multi-turn conversational data proves difficult due to the challenges of user simulation \cite{mehri2026goalalignmentllmbaseduser} and the expensive nature of large-scale manual collection of user interactions \cite{ye2022multiwoz24multidomaintaskoriented}.

While gaps in multi-turn conversational ability have been documented \cite{laban2025llmslostmultiturnconversation, sirdeshmukh2025multichallengerealisticmultiturnconversation}, current measurements of this gap are incomplete and insufficiently informative. Existing multi-turn dialogue benchmarks (e.g., MTBench~\citep{mtbench}, MTEval~\citep{mteval}) are being saturated,
making it difficult to assess what remains of the multi-/single-turn performance gap. Additionally, many conversational benchmarks do not distinguish whether low model performance in multi-turn conversations stems from multi-turn specific performance gaps or from general model ability for similar tasks \cite{mtbench, mtbench101}.

\begin{figure}[t]
    \centering
    \includegraphics[width=\linewidth]{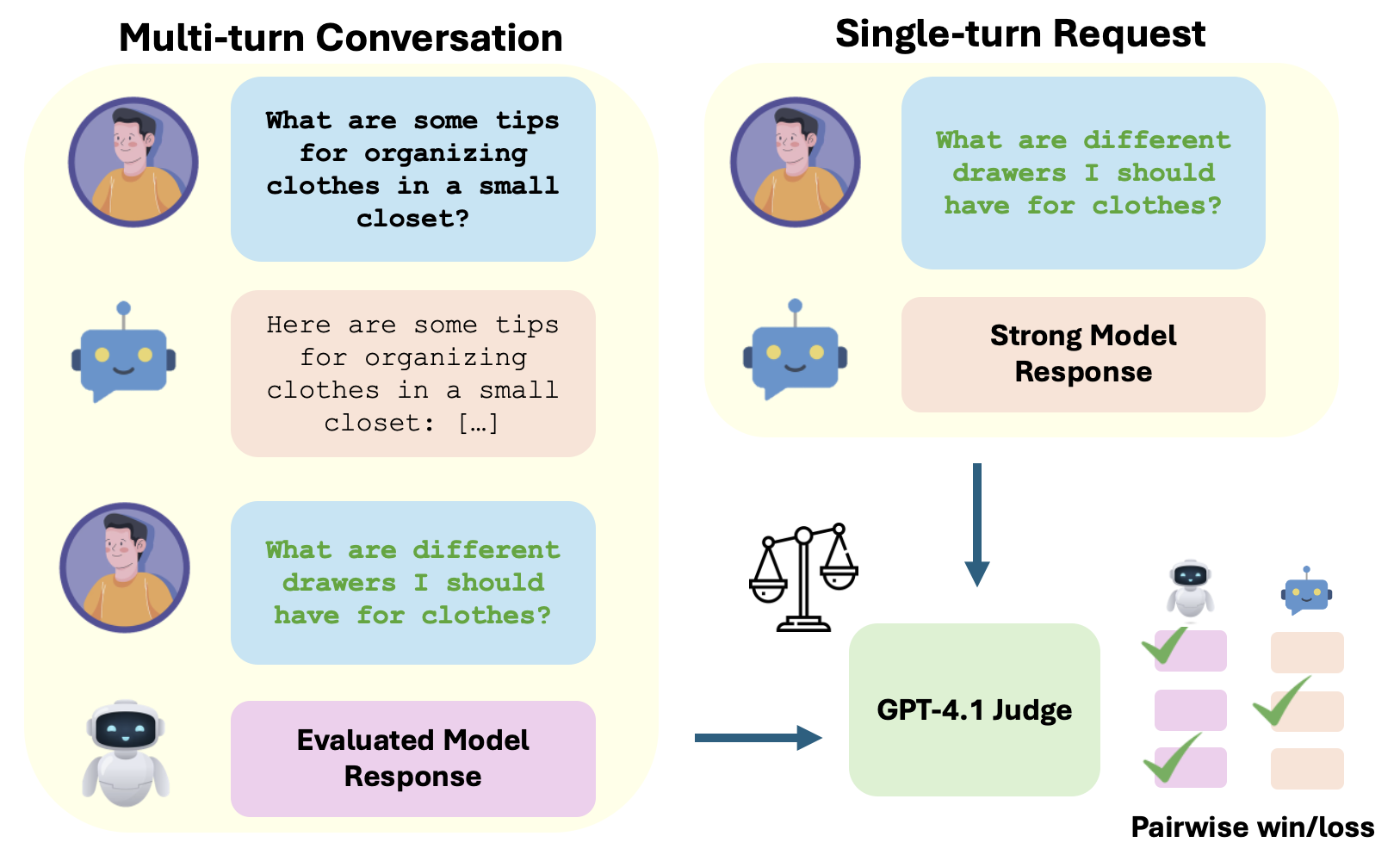}
    \caption{\datasetEval-Absolute example. Model responses to a multi-turn conversation are compared to
    a strong model's response on an equivalent single-turn prompt.
    Our evaluation metric is the win-rate of the evaluated model as determined by an LM judge.}
    \label{fig:turnwiseeval}
\end{figure}

To improve evaluation of the multi-turn conversational abilities of LLMs,
we first introduce a new benchmark, \datasetEval (Section \ref{sec:eval}), which isolates multi-turn abilities from other confounding signals.
Unlike previous multi-turn evaluations, our approach grounds evaluation in pairwise comparisons between responses to multi-turn and equivalent single-turn conversations (Figure \ref{fig:turnwiseeval}).
This methodology isolates multi-turn performance beyond task difficulty, minimizing sensitivity to other LLM capabilities such as factuality and knowledge and providing a more focused assessment of conversational abilities.
Additionally, we use controlled conversational context to further separate multi-turn ability from dependencies on conversational state and to enable better comparisons of results between models.
We show that even frontier models like GPT-5 Chat can underperform on \datasetEval and that open model recipes lag behind.

To address this gap between chat performance of LLMs in single- versus multi-turn conversations, we introduce a scalable method, \dataset, for creating synthetic multi-turn context on arbitrary single-turn conversations (Section \ref{sec:method}). We use data generated with our methodology to analyze how incorporating multi-turn data in training can improve conversational performance with Olmo 3 \cite{olmo3} (Section \ref{sec:exps}).
Through these experiments, we show that adding even a small number of synthetic multi-turn conversations in training can significantly reduce the gap with single-turn performance by up to 12\%, demonstrating the importance of multi-turn conversations in training.

\section{The \datasetEval~Benchmark}
\label{sec:eval}

In this section we detail the motivation, design, and resulting data inside the \datasetEval benchmark. We then show that \datasetEval measures multi-turn conversational abilities for models with a range of single-turn abilities and that multi-turn conversational ability is a distinct dimension of evaluation to consider beyond single-turn chat.

\subsection{Evaluation Design}
To isolate multi-turn conversational ability beyond single-turn task abilities, we use pairwise comparisons of responses to single- and multi-turn conversations.
We consider two comparison settings:

\begin{itemize}
    \item \textbf{\datasetEval-Absolute.} In our first evaluation setting, we measure ``absolute'' multi-turn performance by comparing an evaluated model's completions of multi-turn conversations to responses from a strong model, specifically
    GPT-4 (\texttt{1106-preview}), on equivalent single-turn prompts (Figure \ref{fig:turnwiseeval}). These reference responses represent a strong baseline response to the single-turn setting. This setting uses length-controlled win rate as the primary metric and is intended for easy interpretation of \datasetEval scores across models.

    \item \textbf{\datasetEval-Self.}
    Our second setting further isolates multi-turn ability by comparing an evaluated model's responses to multi-turn conversations to its own responses to corresponding single-turn conversations.
    This setting uses discrete win rate as the primary metric. The single-turn response here represents the model's own baseline ability for the task instruction. Win rates below 50\% imply that a model's responses to multi-turn conversations are consistently worse than their single-turn counterparts, indicating degraded performance in the multi-turn setting beyond what can be attributed to general task ability.
    In best-case scenarios, models can leverage context from previous turns to improve their final response, leading to win rates above 50\%.
\end{itemize}

For pairwise judgments in \datasetEval, we use GPT-4.1 as judge with the judgment prompt from AlpacaEval (Appendix \ref{sec:judge}).

\subsection{Multi-turn Conversation Curation}
Our evaluation captures a common pattern in real-world interactions where users explore related topics across multiple turns.
Our data curation is informed by manual inspection of multiple real-world data sources. To ground our design, we read and analyzed multi-turn interactions from the Ai2 Playground model hosting platform (Appendix \ref{sec:manual}).

\paragraph{Sourcing Prompts.} To build our \datasetEval dataset,
we create multi-turn conversations of up to eight user turns where user turns pose related questions to each seed instruction drawn from AlpacaEval.
Each user turn is synthetically generated and contains a self-contained, topically-related request,
allowing the assistant to optionally leverage earlier turns for richer responses without removing any necessary contextual information from the single-turn comparison conversation.
To control the information seen by the evaluated model in the conversational context so far, we include assistant turns generated by the model used for the reference single-turn completions in \datasetEval-Absolute
to create conversational states for each evaluation sample.
We choose not to go beyond eight user turns to avoid confounding our evaluation with varying long-context abilities\footnote{Conversations to be a maximum of $\sim$4.7k tokens}.
We manually ensured conversational coherence and topical consistency for generated user turns in \datasetEval and verified a random subset of 100 conversations (Appendix \ref{sec:manual}).

\paragraph{Comparison to AlpacaEval.} For reliable comparison to widely-used chat evaluation settings, our final evaluation instructions are sourced from AlpacaEval \cite{alpacaeval} with additional context created as earlier turns in the conversations. Thus, there exists a single-turn complement to each \datasetEval conversation in AlpacaEval, making comparison of these metrics 1:1. This relation with single-turn evaluation allows direct comparison of a model's multi- and single-turn abilities. Comparison between \datasetEval-Absolute and AlpacaEval performance can isolate difference in performance due to the multi-turn setting rather than general model abilities.\footnote{All AlpacaEval results reported here refer to length-controlled AlpacaEval with GPT 4.1 judge (gpt-4.1-2025-04-14). This matches the judge used for \datasetEval and the evaluation setting used by Olmo 3 \cite{olmo3}.}
Specifically, if performance on \datasetEval-Absolute is significantly below that on AlpacaEval, it suggests degraded performance in the multi-turn setting.

\subsection{Demonstrating the Multi-/Single-turn Gap}

\begin{table}[t]
\centering
\resizebox{\linewidth}{!}{
\begin{tabular}{lccc|c}
\toprule
Model & TW-Absolute & AlpacaEval2 & $\Delta$ & TW-Self \\
\midrule
Olmo 3 7B & 36.8 & 42.2 & \textcolor{red}{$-$5.4} & 38.9 \\
Olmo 3.1 32B & 52.4 & 60.1 & \textcolor{red}{$-$7.7} & 34.0 \\
Qwen 3 8B & 48.9 & 49.7 & $-$0.8 & \textbf{53.2} \\
Qwen 3 32B & 67.7 & 69.6 & $-$1.9 & 48.5 \\
Llama 3.1 8B & 18.7 & 16.3 & +2.4 & 40.2 \\
Llama 3.1 70B & 26.3 & 27.7 & $-$1.4 & 36.5 \\
\midrule
GPT-4.1 & \textbf{82.5} & \textbf{83.4} & $-$0.9 & 42.0 \\
GPT-5 Nano & 68.2 & 68.0 & +0.2 & 41.1 \\
GPT-5 Chat & \textbf{83.5} & \textbf{88.5} & \textcolor{red}{$-$5.0} & 40.2 \\
GPT-5.2 & 82.1 & \textbf{83.5} & $-$1.3 & 47.6 \\
\bottomrule
\end{tabular}
}
\caption{Performance of chat models on \datasetEval. Olmo 3 refers to Olmo 3 Instruct models; Qwen 3 results are without thinking enabled. $\Delta$ is the difference between TW-Absolute and AlpacaEval scores.}
\label{tab:turnwiseeval}
\end{table}

Table \ref{tab:turnwiseeval} reports results of different chat models on \datasetEval. Open model recipes (Olmo 3 Instruct 7B and 32B) lag behind on \datasetEval, reflective of the lack of attention multi-turn conversations have received in these efforts.
Meanwhile, even advanced models such as GPT-5 Chat show underperformance on the multi-turn setting compared against their own single-turn abilities.

Additionally, the importance of isolating multi-turn conversational performance from a model's own single-turn baseline is clear; \datasetEval-Self can measure the multi-/single-turn performance gap even when models have particularly low (Llama 3.1) or high (e.g. GPT-4.1) single-turn chat evaluation scores. In these cases, single-turn chat ability becomes the primary influence on all absolute scores, but \datasetEval-Self still reveals models' shortcomings in multi-turn conversations.

\section{Creating Multi-turn Training Data}
\label{sec:method}

We introduce a scalable method for creating synthetic multi-turn data, which we call \dataset. Additionally, we describe a self-talk data baseline which we combine with \dataset conversations in our experimentation in Section \ref{sec:exps}.

\paragraph{Sourcing prompts.}
All conversations build on seed prompts randomly selected from the WildChat subset of the Dolci Instruct SFT mixture,
a primarily single-turn dataset used for Olmo 3 post-training \cite{olmo3}. If seed prompts were selected from already multi-turn conversations, only the first turn was used. We exclude prompts shorter than 15 characters. For both data generation methods, we create conversations with 2 to 8 user turns.

\begin{figure}[t]
    \centering
    \includegraphics[width=\linewidth]{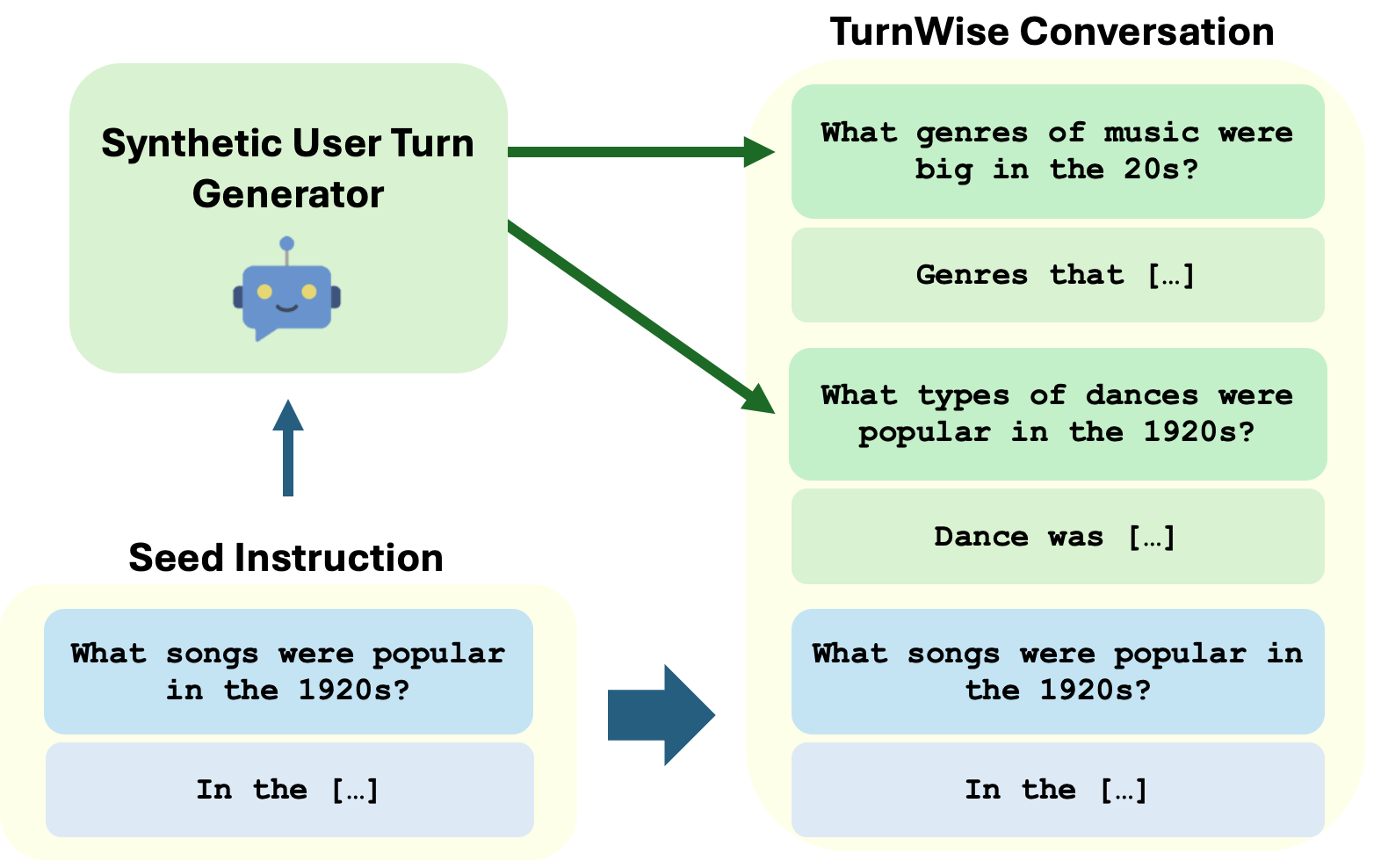}
    \caption{An example of using \dataset to create a synthetic conversation. User turns are independently generated from the seed prompt (green) and conversations are appended with the seed prompt and response (blue).}
    \label{fig:turnwisedata}
\end{figure}

\paragraph{\dataset.}
To create multi-turn conversational data with \dataset, seed prompts are used to independently generate possible user turns. By creating user turns independently, our pipeline avoids issues such as conversational drift, long contexts, and expensive online user simulation. These turns are then stacked to create a synthetic conversation with the original seed prompt as the final turn (Figure \ref{fig:turnwisedata}). We place the seed prompt at the end of the conversation to preserve useful properties of the original dataset such as preference pairs.
User turns were generated with GPT-4.1.

We generate synthetic conversational context from existing prompts with two distinct methods.
Simulating conversations in which users are unsatisfied with an initial response and seek iteration, we generate paraphrased prompts as intermediate user turns.
Similarly, we represent exploratory conversations with follow-up information seeking by generating related queries as intermediate user prompts.

\paragraph{Self-talk}
Prior work has used self-talk as a method for creating artificial conversations where LMs play the part of a user \cite{lambert2024self, Ulmer2024BootstrappingLT}. We use self-talk as a baseline for online conversation generation without user simulation. To capture a diverse range of follow-up interactions, we consider three user behaviors to create user turns: (1) amending previous instructions, (2) seeking clarification on a response, and (3) building on new assistant-provided information.

\section{Training with Multi-turn Data}
\label{sec:exps}

To understand the importance of multi-turn data during training, we conduct experiments with an open-data model, Olmo 3 7B Instruct \cite{olmo3}. We consider two training stages: supervised fine-tuning (SFT) and preference tuning.

\subsection{Experimental Setting}

We create two types of training data: supervised fine-tuning data and preference pairs. In both settings, we create data with \dataset and with self-talk. For creating SFT data, we use GPT 4.1 and Qwen 3 32B (without thinking) to generate assistant turns.
To construct preference pairs, we use the delta-learning heuristic proposed by \citet{geng2025deltalearninghypothesispreference}.
We use two model pairs for data creation: (1) rejected responses produced by GPT-3.5 Turbo and preferred responses by GPT-4.1 and (2) rejected responses by Qwen 3 0.6B and preferred by Qwen 3 32B both without thinking enabled.

For our training baselines, we focus on supervised fine-tuning and Direct Preference Optimization (DPO) \cite{dpo}.
We perform all training experiments on Olmo3-7B-Instruct-SFT. For SFT experiments, we continue training with additional multi-turn SFT data. For preference tuning, we use the full Olmo 3 7B preference mixture augmented with our synthetic data. All experiments use the same hyperparameters and training libraries as in Olmo 3 \cite{olmo3}.

\subsection{Effectiveness of Multi-turn Data}

\begin{table*}[t]
\centering
\resizebox{\textwidth}{!}{
\begin{tabular}{lcccc|cc}
\toprule
Training Data & TW-Absolute & AlpacaEval2 & $\Delta$ & TW-Self & IFEval & MMLU \\
\midrule
SFT (original) & 10.0 & 23.9 & \textcolor{red}{$-$13.9} & 30.1 & \textbf{81.3} & \textbf{66.7} \\
+ 10k \dataset conversations & \textbf{24.1} & 25.1 & $-$1.0 & \textbf{42.9} & 76.7 & 61.4 \\
+ 10k \dataset and self-talk & 20.7 & \textbf{25.6} & \textcolor{red}{$-$4.9} & 41.9 & 72.8 & 63.2 \\
+ 20k \dataset and self-talk & 22.2 & 24.8 & $-$2.6 & 42.7 & 75.6 & 62.7 \\
\midrule
DPO (no multi-turn) & 37.4 & 42.8 & \textcolor{red}{$-$5.4} & 35.0 & 82.6 & 68.0 \\
+ 10k \dataset conversations & 39.1 & 43.8 & \textcolor{red}{$-$4.7} & 41.2 & 81.5 & \textbf{68.3} \\
+ 10k \dataset and self-talk & 38.7 & \textbf{44.9} & \textcolor{red}{$-$6.2} & 40.9 & \textbf{83.2} & 68.1 \\
+ 20k \dataset and self-talk & \textbf{43.4} & 44.7 & $-$1.3 & \textbf{44.2} & 82.8 & 68.0 \\
\bottomrule
\end{tabular}
}
\caption{Performance of models trained with \dataset conversations. All training is done on Olmo 3 7B Instruct SFT models. The SFT (original) line represents the base SFT model and each additional line represents fine-tuning with different additional multi-turn data. The DPO (no multi-turn) line represents the SFT model preference tuned with only single-turn Olmo 3 preference data; each additional line represents adding different additional data to the preference mixture. $\Delta$ = \datasetEval-Absolute $-$ AlpacaEval.}
\label{tab:combined}
\end{table*}

Table \ref{tab:combined} shows that the single- vs. multi-turn performance gap is narrowed by fine-tuning or preference tuning with synthetic multi-turn data (see Table \ref{tab:combined}). Training with \dataset conversations improves the \datasetEval-Self score by up to 12.8 points with fine-tuning and 9.2 points in preference-tuning. Notably, improvements can arise even when synthetic multi-turn conversations comprise less than 5\% of the preference mixture.

\paragraph{Effect on Single-Turn Baselines.}

Table \ref{tab:combined} additionally reports the effect of training with our multi-turn data on single-turn instruction-following \cite{ifeval} and knowledge \cite{mmlu} baselines. Single-turn chat capabilities remain primarily unaffected by the addition of multi-turn data during preference-tuning.
This stability allows us to achieve improvements on \datasetEval while maintaining single-turn performance.
However, supervised fine-tuning with synthetic multi-turn data did decrease IFEval and MMLU performance.
We conjecture that this is due to higher sensitivity to the data generator's abilities with SFT than with preference tuning since preference tuning is less dependent on the absolute quality of the data. Thus, we recommend using preference-tuning or strong data generators to avoid single-turn performance degradation.

Additionally, multi-turn tuned models showed some improvements on (single-turn) AlpacaEval. We conjecture that this is a result of introducing more general chat data.
The effect of introducing more chat data is further explored in Appendix \ref{sec:turns}.

\paragraph{\dataset Effectiveness}
To isolate the effectiveness of \dataset conversations from that of self-talk data, we additionally perform experiments with only \dataset conversations. With the same amount of additional data, fine-tuning achieved better results with \dataset conversations than with the self-talk/\dataset mixture (Table \ref{tab:combined}) implying stronger data from \dataset than the self-talk baseline.
Moreover, training with more multi-turn data leads to stronger performance.

\subsection{Effect of Previous Conversational Turns}
Recent work has shown how earlier turns in a conversation can shift final responses to either improve \cite{Javaji2025AnotherTB} or degrade \cite{laban2025llmslostmultiturnconversation} performance. We control for conversational state in \datasetEval by generating intermediate assistant turns with a strong model to isolate final turn performance and decrease noise sources. To understand how this affects scores,
we experiment with the effect of intermediate response quality in Appendix \ref{sec:weak_context}.

\begin{figure}[h]
    \centering
    \includegraphics[width=0.9\linewidth]{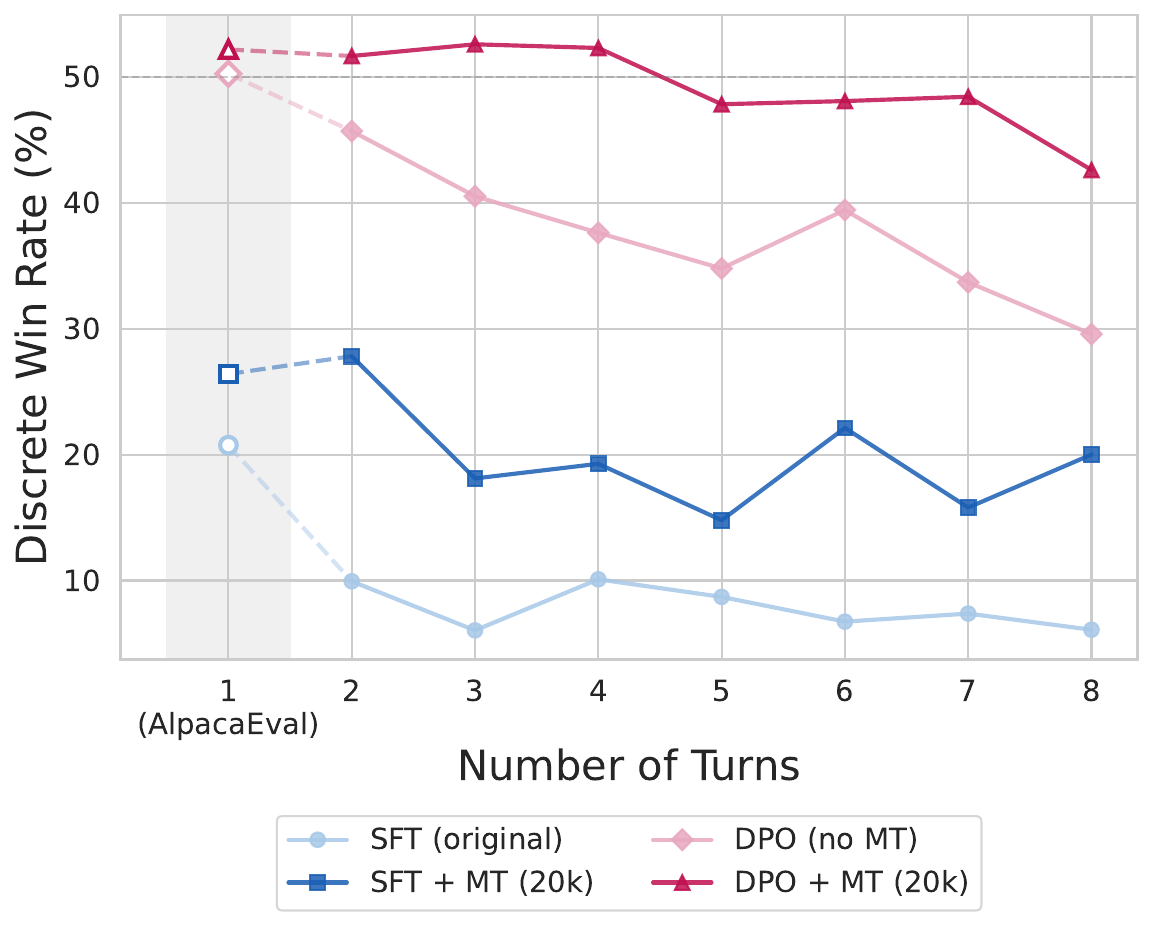}
    \caption{Win rate of models on \datasetEval-Absolute decays across conversation turns. One-turn conversations are represented by AlpacaEval.}
    \label{fig:turns}
\end{figure}

Additionally, we examined performance patterns across conversational lengths to understand how greater conversational context affects model performance (Figure \ref{fig:turns}). Results show that
training with multi-turn data significantly increases performance at every conversational turn and flattens the decay of performance through longer conversations.

\section{Related Work}

\paragraph{Synthetic Multi-turn Data.}
Prior work has considered creation of synthetic multi-turn data. SDSD~\cite{lambert2024self}, for example, creates multi-turn data through LMs talking to themselves, an approach called ``self-talk'' \cite{Ulmer2024BootstrappingLT}.
Other efforts focus on user simulation to create realistic synthetic user turns \cite{prabhakar2025apigenmtagenticpipelinemultiturn, shim2025tooldialmultiturndialoguegeneration, niu-etal-2024-enhancing}. However, user simulations introduce their own challenges such as dialogue state tracking \cite{niu-etal-2024-enhancing} and goal-alignment \cite{mehri2026goalalignmentllmbaseduser}.

\paragraph{Multi-turn Evaluations.}
Prior evaluations of multi-turn capabilities in LLMs have introduced various core skills that arise in multi-turn settings \cite{mteval, mtbench101}. For example, MT-Bench-101 establishes a three-tiered taxonomy founded in educational psychology \cite{mtbench101}. To address saturation of multi-turn benchmarks, MultiChallenge combines hard tasks in long conversations \cite{sirdeshmukh2025multichallengerealisticmultiturnconversation}.

\datasetEval differs from previous multi-turn benchmarks by anchoring judgments in pairwise comparisons to the single-turn setting.
While prior multi-turn benchmarks have used pairwise evaluation \cite{mtbench}, they do not isolate the difference between multi- and single-turn performance. Our comparison of these settings uniquely identifies how performance in multi-turn settings differs from that of single turns.

\section{Conclusion}

To address the single-/multi-turn data gap, we introduce \dataset, a scalable method for generating multi-turn conversations. Additionally, we introduce a new multi-turn benchmark, \datasetEval, to evaluate the difference in a model's performance between single- and multi-turn settings.
Our experiments with training on synthetic multi-turn conversations demonstrate significant improvements to addressing this gap in open model recipes.
Our results suggest that multi-turn conversational ability represents a distinct dimension of model ability not captured by traditional single-turn evaluation, and we recommend greater attention to multi-turn conversations in open data pipelines.

\section*{Limitations}
\datasetEval scores may vary with the quality of the intermediate turns, as analyzed by using a weaker model to generate the responses in context (Table \ref{tab:weak_context}). While this affects absolute score, it does not typically affect overall model rankings. Relative changes in scores reflect evaluated models' robustness to lower quality previous turns in context.

Additionally, our experimental settings are limited in scope to focus on dimensions specific to multi-turn training. Thus, we fix training hyperparameters to those of the Olmo 3 7B Instruct model since the Olmo 3 Instruct data comprises the vast majority of the training data seen by these models.

\bibliography{custom}

\clearpage

\appendix

\section{Judge Template}
\label{sec:judge}
See Figure \ref{fig:judge} for the pairwise judgment template. We use the same judgment template provided by the AlpacaEval repository\footnote{\url{https://github.com/tatsu-lab/alpaca_eval/blob/cd543a149df89434d8a54582c0151c0b945c3d20/src/alpaca_eval/evaluators_configs/alpaca_eval_clf_gpt4_turbo/alpaca_eval_clf.txt}}. All reported evaluations are from individual runs.

\begin{figure*}[h]
\begin{tcolorbox}[colback=yellow!6!white,colframe=blue!50!green]
<|im\_start|>system\\
You are a highly efficient assistant, who evaluates and selects the best large language model (LLMs) based on the quality of their responses to a given instruction. This process will be used to create a leaderboard reflecting the most accurate and human-preferred answers.\\
<|im\_end|>\\
<|im\_start|>user\\
I require a leaderboard for various large language models. I'll provide you with prompts given to these models and their corresponding outputs. Your task is to assess these responses, and select the model that produces the best output from a human perspective.\\

\#\# Instruction

\{\\
\hspace*{10pt} "instruction": """\{instruction\}""",\\
\}\\

\#\# Model Outputs

Here are the unordered outputs from the models. Each output is associated with a specific model, identified by a unique model identifier.

\{\\
\hspace*{10pt}    \{\\
\hspace*{20pt}        "model\_identifier": "m",\\
\hspace*{20pt}        "output": """\{output\_1\}"""\\
\hspace*{10pt}    \},\\
\hspace*{10pt}    \{\\
\hspace*{20pt}        "model\_identifier": "M",\\
\hspace*{20pt}        "output": """\{output\_2\}"""\\
\hspace*{10pt}    \}\\
\}\\

\#\# Task

Evaluate the models based on the quality and relevance of their outputs, and select the model that generated the best output. Answer by providing the model identifier of the best model. We will use your output as the name of the best model, so make sure your output only contains one of the following model identifiers and nothing else (no quotes, no spaces, no new lines, ...): m or M.\\

\#\# Best Model Identifier\\
<|im\_end|>
\end{tcolorbox}
\caption{The judge template for pairwise comparison.}
\label{fig:judge}
\end{figure*}

\begin{figure*}[h]
\begin{tcolorbox}[colback=yellow!6!white,colframe=blue!50!green]
\textbf{User}: What are some memorable game-winning plays in sports history?\\\\
\textbf{Assistant}: There have been many unforgettable game-winning plays in sports history that have left fans in awe. Some of the most iconic ones include Michael Jordan's last shot with the Chicago Bulls in the 1998 NBA Finals, known as "The Last Shot," where he hit a jumper to secure the championship. Another legendary play is David Tyree's helmet catch in Super Bowl XLII, where he pinned the ball against his helmet to help the New York Giants defeat the New England Patriots. And who can forget the "Miracle on Ice" in the 1980 Winter Olympics when the USA hockey team upset the Soviet Union in a thrilling victory. These moments showcase the magic and drama that sports can bring.\\\\
\textbf{User}: As a sports commentator, describe the winning play in the final seconds of a championship game.\\\\
\textbf{Assistant}:

\end{tcolorbox}
\caption{An example conversation from \datasetEval.}
\label{fig:task}
\end{figure*}

\section{Manual Inspection of Data}
\label{sec:manual}

Our benchmark creation is guided by inspection of over 200 conversations with T\"ulu 3 collected from the Ai2 model hosting interface. We additionally inspected multi-turn conversations from WildChat and LMSYS Chatbot Arena to verify our observed behaviors. From our inspection, we observed two common modes of user behaviors. First, users enter the conversation with a specific target and continue the conversation through multiple turns to iterate on the output. Second, users enter exploratory conversations that included following-up on information from previous turns in the conversation. Our synthetic data methodology is grounded in these observed conversation types.

\paragraph{Manual verification of data quality} Figure \ref{fig:task} shows an example of a conversation from \datasetEval. To ensure a high-quality evaluation, all conversations in \datasetEval were tagged by GPT 4.1 for coherence and degenerate turns. A total of 64 turns across 38 conversations were flagged for quality. Each flagged turn was manually inspected, and where relevant, the user turn was rewritten (and response regenerated) or the turn was removed. As a result, 22 turns were edited and 39 turns removed across 17 conversations. Final verification of a random sample of 100 conversations showed topical relevance and coherent instructions across all turns.

\section{Turn Number Ablations}
\label{sec:turns}
We perform additional training ablations using truncated versions of the augmented multi-turn conversations (Table \ref{tab:turns}). Our results demonstrate that improvement goes beyond benefits from training with more chat data as substituting our synthetic multi-turn data with truncated single-turn versions cannot account for all improvements in \datasetEval.

\begin{table}[h]
\centering
\resizebox{\linewidth}{!}{
\begin{tabular}{lcccc|cc}
\toprule
Model & TW-A & AE2 & $\Delta$ & TW-S & IFEval & MMLU \\
\midrule
SFT (original) & 10.0 & 23.9 & \textcolor{red}{$-$13.9} & 30.1 & \textbf{81.3} & \textbf{66.7} \\
+ max 1 turn & 16.9 & 25.5 & \textcolor{red}{$-$8.6} & 40.2 & 73.0 & 59.7 \\
+ max 2 turn & 20.7 & 22.7 & $-$2.0 & \textbf{43.5} & 76.3 & 62.2 \\
+ max 4 turn & \textbf{21.3} & 25.3 & \textcolor{red}{$-$4.0} & 41.8 & 75.6 & 63.1 \\
+ max 8 turn & 20.7 & \textbf{25.6} & \textcolor{red}{$-$4.9} & 41.9 & 72.8 & 63.2 \\
\midrule
DPO (no MT) & 37.4 & 42.8 & \textcolor{red}{$-$5.4} & 35.0 & 82.6 & 68.0 \\
+ max 1 turn & 38.6 & \textbf{50.1} & \textcolor{red}{$-$11.5} & 38.9 & 83.4 & 68.0 \\
+ max 2 turn & 38.1 & 45.8 & \textcolor{red}{$-$7.7} & 41.6 & 82.1 & \textbf{68.5} \\
+ max 4 turn & \textbf{39.3} & 45.0 & \textcolor{red}{$-$5.7} & \textbf{42.9} & \textbf{83.7} & 68.2 \\
+ max 8 turn & 38.7 & 44.9 & \textcolor{red}{$-$6.2} & 40.9 & 83.2 & 68.1 \\
\bottomrule
\end{tabular}
}
\caption{Ablations of long conversations in multi-turn training data. + max $n$ turns indicates truncating the additional multi-turn data to $n$ user turns.}
\label{tab:turns}
\end{table}

\section{Effect of Context Response Quality.}
\label{sec:weak_context}
We hypothesize that high-quality intermediate responses in context can inform and improve the final evaluated responses. To investigate this potential for in-context learning, we examine how intermediate response quality affects performance. We created an alternative version of \datasetEval using GPT-3.5 Turbo generated responses for this analysis.
Results show that scores with this alternative dataset preserve rankings and conclusions about model abilities (Table \ref{tab:weak_context}). This suggests that a model's success on \datasetEval is not solely driven by high-quality in-context examples.

\begin{table}[h]
\centering
\resizebox{\linewidth}{!}{
\begin{tabular}{lcccc}
\toprule
Model & TW-A & TW-A (Weak) & TW-S & TW-S (Weak) \\
\midrule
SFT (original) & 10.0 & 9.4 & 30.1 & 28.4 \\
+ MT (10k) & 20.7 & 20.4 & 41.9 & 40.9 \\
+ MT (20k) & \textbf{22.2} & \textbf{23.3} & \textbf{42.7} & \textbf{43.0} \\
\midrule
DPO (no MT) & 37.4 & 35.7 & 35.0 & 33.0 \\
+ MT (10k) & 38.7 & 40.3 & 40.9 & 39.4 \\
+ MT (20k) & \textbf{43.4} & \textbf{44.3} & \textbf{44.2} & \textbf{42.7} \\
\bottomrule
\end{tabular}
}
\caption{Performance of models trained with synthetic multi-turn data on \datasetEval with weak context. TW-A (Weak) and TW-S (Weak) indicate alternative versions of \datasetEval that use GPT-3.5 Turbo generated assistant turns in context.  + MT indicates training with additional multi-turn data (\dataset and self-talk). }
\label{tab:weak_context}
\end{table}

\end{document}